\documentclass[]{spie} 
\usepackage[]{graphicx}
\usepackage{epstopdf}
\usepackage{float}
\usepackage{bm}
\usepackage{amsmath}
\usepackage{amsfonts}
\usepackage{amssymb}
\usepackage{algorithmic}
\usepackage{algorithm}
\usepackage{xcolor}

\DeclareMathOperator*{\argmin}{arg\,min}

\title{Buried object detection using handheld WEMI with task-driven extended functions of multiple instances}

\author{Matthew Cook, Alina Zare, Dominic K. C. Ho
	\skiplinehalf
	University of Missouri, Columbia, MO 65211
}

\begin{document}

	\maketitle

	\begin{abstract}
		Many effective supervised discriminative dictionary learning methods have been developed in the literature.  However, when training these algorithms, precise ground-truth of the training data is required to provide very accurate point-wise labels.  Yet, in many applications, accurate labels are not always feasible. This is especially true in the case of buried object detection in which the size of the objects are not consistent.  In this paper, a new multiple instance dictionary learning algorithm for detecting buried objects using a handheld WEMI sensor is detailed.  The new algorithm, Task Driven Extended Functions of Multiple Instances, can overcome data that does not have very precise point-wise labels and still learn a highly discriminative dictionary.  Results are presented and discussed on measured WEMI data.
	\end{abstract}


	\section{Introduction}
		\label{Intro}
		
		Electromagnetic induction (EMI) sensors have been studied extensively for their use in detecting buried landmines \cite{BEMISDBL, WMMSSMPMO}.  The traditional EMI sensor sends a time-varying electromagnetic field down through the ground to interact with subsurface objects.  When these fields interact with buried objects they induce a response that can then be measured by the sensor, indicating there is some metal content in the buried object \cite{WEMIPLD}.
		
		The data used in this investigation was collected utilizing a handheld sensor system that contained both an EMI sensor and ground penetrating radar that work together to detect buried objects.  The two sensors function independently of each other to enable specialized algorithms for each sensor whose outputs can then be fused to create a final detection measure.  In this paper, an algorithm for the detection of buried objects using the handheld EMI sensor is presented.
		
		Previous work has explored the use of data modeling and fixed dictionaries for detection.  Several algorithms have set out to model the metallic objects such as the Gradient Angle Model Algorithm \cite{GRANMA} which computes properties of metallic objects and then searches for these in the data.   Another approach to model the signature of metallic objects is the Discrete Spectrum of Relaxation Frequencies (DSRF) \cite{EMIDSRF} which seeks to build a model of metallic objects in the frequency domain.
		
		Other methods such as the Joint Orthogonal Matching Pursuits (JOMP) \cite{LDTTJOMP} use a fixed dictionary generated from the DSRF model to look for potential landmines.  The JOMP algorithm also returns which dictionary element was chosen for a particular data point which allows other algorithms to be trained on this output and improve the results even further.  The Possibilistic $K$-Nearest Neighbors (PKNN) \cite{LCPKNNWEMI} algorithm is one such algorithm that can then classify the JOMP output to determine the type of object that is found by JOMP.
		
		More recently, however, multiple instance dictionary learning algorithms have been created to learn a dictionary from imprecisely labeled data \cite{MIDLSOD}.  This paper proposes a new multiple instance dictionary learning algorithm that builds upon the earlier work.
		
		In Section \ref{Bgrnd}, several algorithms related to the proposed algorithm are briefly reviewed. In Section \ref{TDFM}, the Task Driven Extended Functions of Multiple Instances (TD-$e$FUMI) algorithm is introduced.  Section \ref{Res} presents TD-$e$FUMI results on an EMI data set.  Finally, Section \ref{Summ} provides a discussion of future work.
		
	\section{Background}
		\label{Bgrnd}
		
		Before moving onto the specifics of the TD-$e$FUMI algorithm, first a review is provided of some of the previously published methods upon which the TD-$e$FUMI algorithm extends. 
		
		\subsection{Discrete Spectrum of Relaxation Frequencies}
			\label{DSRF}
						
			The Discrete Spectrum of Relaxation Frequencies (DSRF) \cite{REDSREIR} is a model that represents buried metallic objects in wide-band EMI data.  The DSRF is a discrete case of the Distribution of Relaxtion Times (DRT) model \cite{NRMCRS}.  The DSRF model is formulated in terms of relaxation frequencies (where, in contrast, the DRT is in terms of relaxation times).  The benefits to using the DSRF is that it is directly related to the physical properties of potential targets and is invariant to the relative position and orientation of the target to the EMI sensor\cite{REDSREIR}.
			
			The DSRF model is as follows,			
			\begin{equation}
				H(\omega) = c_0+\sum\limits_{k=1}^K\frac{c_k}{1+j\omega/\zeta_k},
				\label{DSRFeqn}
			\end{equation}
		where $c_0$ is the DC shift, $c_k$ is the real spectral amplitudes at each of the $K$ relaxation frequencies $\zeta_k$ (relaxation frequencies are in radians), and $\omega$ is the transmitted frequency in radians.   In previous work, several detection algorithms have leveraged the DSRF model to generate features that can then be used to detect landmines \cite{LDUDSRF, LDTTJOMP, EADSREMIR}.
			
\begin{figure}[thb]
			\centering
			\includegraphics[width=.6\textwidth, trim={0 0.55cm 0 0cm}, clip]{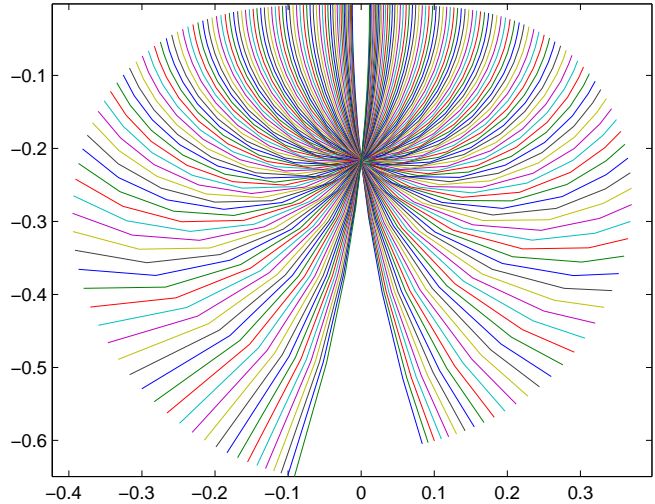}
			\caption{DSRF dictionary created from Equation \ref{DSRFdict}.}
			\label{DSRF_D}
		\end{figure}

			The DSRF model can also be used to generate representative signals that will model different types of signals.  When a group of these signals are joined they create a dictionary that can also be used for detection.  Since the transmit frequencies, $\omega$, are fixed the dictionary can be generated by simply selecting several different values of $\zeta_k$ then computing the resulting signals.  Generation of the dictionary elements then takes the form,
			\begin{equation}
				\bm{d}_i = \left[\frac{1}{1+j\omega_1/\zeta_i}, \frac{1}{1+j\omega_2/\zeta_i}, \dots, \frac{1}{1+j\omega_N/\zeta_i}\right]^T,
				\label{DSRFdict}
			\end{equation}
			where the values of $\omega$ are known from the sensor.  A dictionary generated via this method is shown in Figure \ref{DSRF_D}.  Each of the created dictionary elements can then be compared to EMI data to help find metal objects.  Since the model only accounts for metal objects the dictionary will, hopefully, only look like the data in locations where metal targets exist. The many dictionary elements will allow for the detection of a larger amount of possible mine-like objects. 
			
			With this fixed dictionary, algorithms can be developed to find these signals in the data.  One such algorithm is the Joint Orthogonal Matching Pursuits (JOMP) \cite{LDTTJOMP}.  JOMP is an algorithm that is developed from the Matching Pursuits (MP) \cite{MPTFD} algorithm which is use to deconstruct a given signal using a known dictionary.  MP is a greedy method that compares each dictionary element to the given signal to find the dictionary element that matches the input signal the closest.  Once the match has been made the dictionary element is subtracted from the given signal.  This process is repeated, with each successive match being made to what is left of the original signal, until the signal becomes negligible, or a set number of dictionary elements have been used.  A slight modification to the MP algorithm is the Orthogonal Matching Pursuits (OMP) \cite{OMPSSRWN} algorithm that behaves the same except that an orthogonal transformation is used to ensure that each new element is orthogonal to each of the previously added dictionary elements.
			
			The JOMP algorithm uses the same methodology as OMP except that instead of deconstructing a single data points with a given dictionary, the JOMP algorithm considers several points simultaneously.  In this implementation of JOMP, two data points are processed simultaneously (one positioned 5 points ahead and one 5 points behind the current location being processed).  JOMP is only selecting one dictionary element to model both points.  JOMP then assigns a confidence based on the average of the inverse of the two residual errors.  The residual error is the error in reconstruction from the chosen dictionary element.
			
			In some cases, results can be improved by learning a dictionary from the data as opposed to using a fixed universal dictionary.  Thus, in this paper, we develop an approach to estimate a dictionary that optimizes target detection results (with respect to an objective function as defined in the following section). Our approach for estimating this dictionary from the data is an extension of Task-driven Dictionary Learning\cite{TDDL} and the Extended Functions of Multiple Instances\cite{EFMITC} approaches. 

		\subsection{Task-Driven Dictionary Learning}
			\label{TDDL}
			
			The Task-Driven Dictionary Learning (TDDL) \cite{TDDL} algorithm is a supervised dictionary learning model that simultaneously learns a classifier with the dictionary to aid classification results.  The algorithm uses a two layer model, the top layer being the classification error, 			
			\begin{equation}
				\min_{\bm{D}\in \mathcal{D}, \bm{w}\in \mathcal{W}} {\rm I\!E}_{y,x}[\ell_s(y,\bm{w},\bm{\alpha}^*(\bm{x},\bm{D}))] + \frac{v}{2}||\bm{w}||_F^2,
				\label{OBJ}
			\end{equation}
			where $\ell_s$ is a loss function comparing the actual class label $y$ to the estimated class label computed from the classifier $\bm{w}$ and the sparse weights $\bm{\alpha}^*(\bm{x},\bm{D})$ which is a function of the known dictionary $\bm{D}$ and the current data point $\bm{x}$.  This model uses a simple linear classifier to do the classification,
			\begin{equation}
				\hat{y} = \bm{w}^\intercal \bm{\alpha}^*+\psi,
				\label{TDDL_out}
			\end{equation}
			where $\hat{y}$ is the estimated class label and $\psi$ is an offset or threshold.  In Equation \ref{OBJ} the loss function can be replaced with any twice differentiable convex loss function.	
			
			The second layer of the TDDL model is the dictionary learning/sparse coding layer.  In this layer, the objective function used to determine the dictionary and sparse weights is:
			\begin{equation}
				\bm{\alpha}^*(\bm{x},\bm{D}) \stackrel{\Delta}{=} \argmin_{\bm{\alpha} \in {\rm I\!R}^{K}} \frac{1}{2}||\bm{x}-\bm{D\alpha}||_2^2+\lambda_1||\bm{\alpha}||_1+\frac{\lambda_2}{2}||\bm{\alpha}||_2^2.
				\label{E-net}
			\end{equation}
			In this model the dictionary is also subject to the constraint:			
			\begin{equation}
				\mathcal{D}\stackrel{\Delta}{=}\{D\in{\rm I\!R}^{L\times K}\text{s.t.}\forall k\in\{1,\dots,K\},||\bm{d}_k||_2\leq 1\},
				\label{constr}
			\end{equation}
			which limits each dictionary element to be on or inside of the unit hypersphere.
			
			The optimization for this algorithm is done using stochastic gradient descent \cite{LSMLSGD} with a mini-batch strategy, i.e. a small number of data points are used to complete each update instead of a single data point.  In each step of the optimization process includes two sub-steps: the first sub-step is to compute the sparse weights, then, the second sub-step is to use these sparse weights to compute updates for both the classifier $\bm{w}$ and the dictionary $\bm{D}$.
			
			This algorithm will learn a discriminative dictionary trained for the target detection task and, furthermore, trains a classifier that can be used for target detection.  However, this algorithm also introduces a new potential problem, namely, as TDDL is a supervised learning algorithm it requires precisely labeled ground-truth data (i.e., ground-truth with data point level accuracy). Obtaining precisely labeled data can be very difficult to acquire for buried objects.

		\subsection{Extended Functions of Multiple Instances}
			\label{eFM}
			
			The Extended Functions of Multiple Instances algorithm ($e$FUMI)\cite{EFMITC} is a form of multiple instance dictionary learning that separates the target and non-target dictionary elements.  This algorithm has a built in method to estimate whether any given point contains any portion of target.  This estimation allows the algorithm to determine if a point that is labeled as target actually contains any portion of the target dictionary elements.  The structure of this algorithm is 			
			\begin{equation}
				\begin{aligned}
					F = \frac{-(1-u)}{2}\sum\limits_{n=1}^N & \left\| \bm{x}_n - z_n\sum\limits_{t=1}^T\alpha_{nt}\bm{d}^+_t - \sum\limits_{m=1}^M\alpha_{nm}\bm{d}^-_m\right\|_2^2\\ 
					&- \frac{u}{2}\left( \sum_{m=1}^M\left\|\bm{d}^-_m-\bm{\mu}_0\right\|_2^2 + \sum\limits_{t=1}^T\left\|\bm{d}^+_t-\bm{\mu}_0\right\|_2^2 \right) - \sum_{m=1}^M\gamma_m\sum_{n=1}^N\alpha_{nm},
				\end{aligned}
				\label{eFUMI_eqn}
			\end{equation}
			where $M$ is the number of non-target dictionary elements, $T$ indicates the target dictionary element/weight, $\bm{\mu}_0$ is the global data mean, $\gamma_m$ is a sparsity promoting term on the number of dictionary elements, $z_n$ is an unknown latent variable, and the remaining terms are carried over from Equations \ref{OBJ} and \ref{E-net}.
			
			The latent variable $z_n$ is an indicator variable that tells the algorithm whether the current data point is a true target or not.  The indicator is then used to determine whether the full dictionary should be used for reconstruction or just the non-target dictionary.  In order to minimize the objective function when this value is unknown (as is the case for positively-labeled points under a multiple instance learning framework), expectation maximization (EM) is used to find the expected value of $z_n$ for each data point.  When evaluating the expectation of Equation \ref{eFUMI_eqn} with respect to $z_n$ the resulting expression is			
			\begin{equation}
				\begin{aligned}
					\mathbb{E}[F] = \sum_{\substack{z_n\in\{0,1\}}} \left[ -\frac{1}{2}(1-u)\sum_{n=1}^N  P(z_n|\mathbf{x}_n, \boldsymbol{\theta}^{(i-1)})\left\| \mathbf{x}_n - z_n\alpha_{nT}\mathbf{d}_T - \sum_{m=1}^M\alpha_{nm}\mathbf{d}_m\right\|_2^2\right]\\ 
					- \frac{u}{2}\sum_{m=1}^M\left\|\mathbf{d}_m-\boldsymbol{\mu}_0\right\|_2^2 - \frac{u}{2}\left\|\mathbf{d}_T-\boldsymbol{\mu}_0\right\|_2^2 - \sum_{m=1}^M\gamma_m\sum_{n=1}^N\alpha_{nm} ,
					\label{EM_eFUMI}
				\end{aligned}
			\end{equation}
			where the probabilities, $P(z_n|\mathbf{x}_n, \theta^{(i-1)})$, are computed via
			\begin{equation}
				\begin{aligned}
					P(z_n|&\mathbf{x}_n, \theta^{(i-1)}) = \\
					&\left\{ \begin{array}{l}
					p(z_n=0|\mathbf{x}_n\in B_j^+, \boldsymbol{\theta}^{(i-1)})=\exp\left(-\beta \left\| \mathbf{x}_n - \sum\limits_{m=1}^M\alpha_{nm}\mathbf{d}_m\right\|_2^2\right)\\
					p(z_n=1|\mathbf{x}_n\in B_j^+, \boldsymbol{\theta}^{(i-1)})=1-\exp\left(-\beta \left\| \mathbf{x}_n - \sum\limits_{m=1}^M\alpha_{nm}\mathbf{d}_m\right\|_2^2\right)\\
					p(z_n=0|\mathbf{x}_n\in B_j^-, \boldsymbol{\theta}^{(i-1)})=1\\
					p(z_n=1|\mathbf{x}_n\in B_j^-, \boldsymbol{\theta}^{(i-1)})=0
					\end{array} . \right.
					\label{Prob_upd}
				\end{aligned}
			\end{equation}
			The probabilities that are computed are dependent on only how well the non-target dictionary elements can reconstruct the current data point and a scaling parameter $\beta$.  If the non-target dictionary elements can accurately reconstruct the current data point then the data point is assigned a high probability of being non-target, while if the non-target dictionary elements cannot describe the data, a high probability of target is then assigned.  For points labeled as non-target, the probability of target is set to zero for the training because they are known to be non-target.
			
			The $e$FUMI algorithm gives a way to handle imprecise training data but unlike the TDDL algorithm it does not leverage a classifier to learn a discriminative dictionary that aims to boost the classification performance.
		
	\section{Task-Driven Extended Functions of Multiple Instances}
		\label{TDFM}
		
		The proposed Task-Driven Extended Functions of Multiple Instances (TD-$e$FUMI) is a combination of the TDDL and $e$FUMI algorithms. Therefore the TD-$e$FUMI algorithm will be a multiple instance learning algorithm that also learns a classifier alongside the dictionary.  The classification model is similar to the TDDL model while the sparse weight/dictionary model was taken from $e$FUMI.  
		
		The combined classification model is 		
			\begin{equation}
				\begin{aligned}
					\mathcal{F}(\bm{w},\bm{\alpha}_n^*(\bm{x}_n,\bm{D}))=\mathbb{E}_z&\left[\frac{(1-u)\delta_n}{2}\left(z'_n-\bm{w}^\intercal\bm{\alpha}_n^*(\bm{x}_n,\bm{D})\right)^2\right]\\
					& +\frac{u}{2}\left( \sum\limits_{t=1}^T\left\| \bm{d}^{(T)}_t-\bm{\mu}_0 \right\|_2^2+\sum\limits_{m=1}^M\left\| \bm{d}^{(NT)}_m-\bm{\mu}_0 \right\|_2^2 \right)+\frac{v}{2}\left\|\bm{w}\right\|_2^2\\
					& + \sum\limits_{k=1}^K\sum\limits_{l=2}^L\frac{s}{2}\left(\bm{d}_k(l)-\bm{d}_k(l-1)\right)^2 ,
				\end{aligned}
				\label{comp_obj_TDF}
			\end{equation}
			where
			\begin{equation}
				\delta_n = \left\{
				\begin{aligned}
					\epsilon \frac{N_M}{N_T}, \ & \ \text{for target points}\\
					1, \ & \ \text{for background points}
				\end{aligned}
				\right. ,
				\label{ratio_correction}
			\end{equation}
			$\bm{d}_t^{(T)}$ and $\bm{d}_m^{(NT)}$ are respectively elements of the target and non-target dictionaries, $\bm{D}_kf(l)$ is the $l^{th}$ dimension value of the $k^{th}$ dictionary element from the entire dictionary, $z'_n$ is an unknown latent variable, and $s$ is the parameter that controls the smoothing term.
			
			For the dictionary learning portion to keep this similar to the TDDL algorithm the same sparse coding function was chosen,
			\begin{equation}
				\bm{\alpha}^*_n(\bm{x}_n,\bm{D})=\argmin_{\bm{\alpha}} \frac{1}{2}\left\| \bm{x}_n-z_n\sum\limits_{t=1}^T\bm{\alpha}_{nt}\bm{d}^{(T)}_t-\sum\limits_{m=1}^M\bm{\alpha}_{nm}\bm{d}^{(NT)}_m\right\|_2^2 + \lambda\|\bm{\alpha}_n\|_1 ,
				\label{nw_MTF_spcd}
			\end{equation}
			with a few modifications.  First the latent variable $z_n$ was added to allow the method to adapt to and learn which data points are target and which are non-target, also the $\ell_2$ norm was dropped from this equation.
			
			During optimization, first the expectation of equation \ref{comp_obj_TDF} with respect to the unknown variable $z'_n$.  The probabilities for $z'_n$ are identical to those for $z_n$ shown in Equation \ref{Prob_upd}.  Once this is done stochastic gradient descent is used to alternatively update the sparse weights and then the dictionary and classifier until the dictionary and classifier converge.
			
			Once the dictionary and classifier have been trained, the classification algorithm consists of only two steps: (1) First the sparse weights of each data point to be tested are found using
			\begin{equation}
				\bm{\alpha}^*_n(\bm{x}_n,\bm{D})=\argmin_{\bm{\alpha}} \frac{1}{2}\left\| \bm{x}_n-\bm{D\alpha}_n\right\|_2^2 + \lambda\|\bm{\alpha}_n\|_1 ,
				\label{tst_spcd}
			\end{equation}
		which is a small modification of Equation \ref{nw_MTF_spcd} where the latent variable is removed, so that the entire dictionary can be used to reconstruct the entire data; (2)  The second step is to compute the algorithm output which is done via Equation \ref{TDDL_out}.

	\section{Experiments and Results}
		\label{Res}
		
			The data used in this investigation was collected using a handheld EMI sensor swept over several lanes mounted on a cart system.  Buried objects are classified as clutter or high, low, or non metal targets.  Six lanes of data are used in the experiments shown in this paper. The number of objects of each type are shown in Table \ref{Tar_desc}.
			
			\begin{table}[thb]
				\centering
				\caption{Object description for EMI data. Abbreviations, HMT: High-Metal Target, LMT: Low-Metal Target, NMT: Non-Metal Target, CL: Clutter.}
				\begin{tabular}{c c c c c c}
					\hline\hline
					 & \textbf{HMT} & \textbf{LMT} & \textbf{NMT} & \textbf{CL}\\
					\textbf{Lane 1} & 4  & 7  & 0  & 6 \\
					\textbf{Lane 2} & 4  & 10 & 0  & 4 \\
					\textbf{Lane 3} & 4  & 7  & 0  & 8 \\
					\textbf{Lane 4} & 6  & 6  & 3  & 0 \\
					\textbf{Lane 5} & 7  & 5  & 5  & 0 \\
					\textbf{Lane 6} & 6  & 6  & 2  & 3 \\
					\textbf{Totals} & 31 & 41 & 10 & 11 \\
					\hline
				\end{tabular}
				\label{Tar_desc}
			\end{table}

		\subsection{Alarm Generation}
			\label{Ala_Gen}
			This algorithm is implemented as a classifier for the data used in these experiments.  Specifically, in our implementation, the algorithm is not run directly on the entirety of the data. Instead alarms or points of interest are found and then the TD-eFUMI algorithm is used to classify each of the returned alarms as either a mine-like object or non-target.
			
			To generate alarms for the TD-$e$FUMI algorithm, the JOMP detector that was mentioned in Section \ref{DSRF} is used.  The typical output from JOMP is a confidence map that indicates where JOMP believes a target exists.  To generate alarms from this confidence map a threshold is applied to the confidence values so as to only keep the most confidence points.  To then find the alarm locations the Mean Shift \cite{MS} algorithm is used.  This algorithm clusters data and returns the centroid of each cluster, this centroid is used as the alarm location.  
			
			Since one data point for each alarm is not enough to train the algorithm, a group of data points in a 0.25m radius around each alarm location is gathered and included in each alarm to act as the alarm data.  The data that is collected to be used for the alarm data is processed after being collected to ensure some consistency throughout the different lanes.  The process for this is only to subtract the mean of the lane from all of the alarm data.  

The last step in generating the alarms is to assign whether each alarm was a target or a false alarm for training. To do this the ground truth was compared to the alarms returned by JOMP and if an alarm location was within 0.25m of any target ground truth location the alarm was labeled as target. The alarm results from the JOMP prescreener are shown in Table \ref{Ala_desc}, the number of targets and false alarms were not made to be equal it is simply a coincidence that the totals match. In this data that leaves six sets of alarms corresponding to the six lanes in the data set.

			
			\begin{table}[thb]
				\centering
				\caption{Alarms found by the JOMP prescreener over the six testing lanes.}
				\begin{tabular}{c c c c}
					\hline\hline
					 & \textbf{Target Alarms} & \textbf{False Alarms} & \textbf{Total Alarms}\\
					\textbf{Lane 1} & 20  & 29  & 49\\
					\textbf{Lane 2} & 19  & 22  & 41\\
					\textbf{Lane 3} & 18  & 20  & 38\\
					\textbf{Lane 4} & 14  & 17  & 31\\
					\textbf{Lane 5} & 19  & 13  & 32\\
					\textbf{Lane 6} & 19  & 8   & 27\\
					\textbf{Totals} & 109 & 109 & 218\\
					\hline
				\end{tabular}
				\label{Ala_desc}
			\end{table}

		\subsection{Alarm Based TD-$e$FUMI}
			\label{Ala_TDF}			
								
\begin{figure}[thb]
			\centering
			\includegraphics[width=.5\textwidth, trim={1.8cm 1cm 1.1cm 0}, clip]{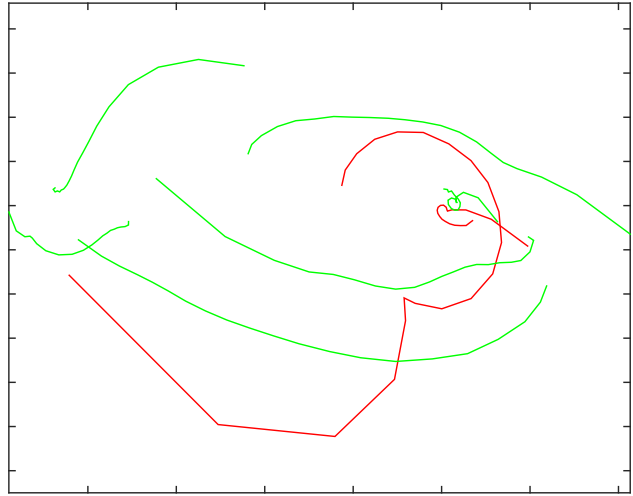}
			\caption{Sample Dictionary learned by TD-$e$FUMI. Target elements are in green and non-target elements in red.}
			\label{D_ex}
		\end{figure}

			To avoid any test on train situations lane based cross validation is used.  This means that when TD-$e$FUMI is being trained only five of the six sets of alarms are used for training data, and the sixth set is used for testing.  As an example when lane one is being tested the alarms from lanes two through six are used to train TD-$e$FUMI.  With this approach TD-$e$FUMI learns six different dictionaries, one for each fold of the cross validation.  An example of a learned dictionary is shown in Figure \ref{D_ex}.
		
		The dictionary learned by the TD-$e$FUMI algorithm here is similar to what is expected in the DSRF model introduced in Section \ref{DSRF}.  The one noticeable difference is that the newly learned dictionary elements are not centered on the real, horizontal, axis.  The reasoning for this is that in the DSRF model the real mean is subtracted from all of the data samples.  In this dictionary only the overall data mean was subtracted, which leaves any shift in the real axis behind.  Leaving the real shifts in the dictionary allows the dictionary to cover a much larger search space during the optimization routine, potentially leading to a better result with this data.
		
\begin{figure}[thb]
			\centering
			\includegraphics[width=.6\textwidth, trim={0 0.55cm 0 0cm}, clip]{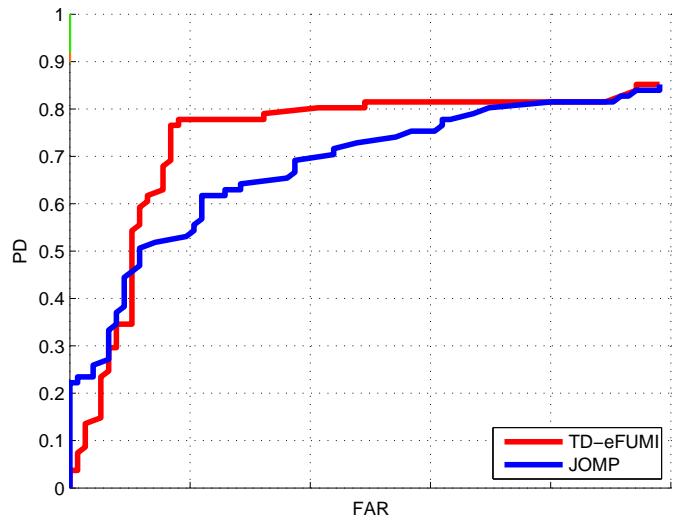}
			\caption{ROC curve comparing the classification performance of TD-$e$FUMI with the prescreener that was used to generate alarms.}
			\label{ROC_JMPvTDF}
		\end{figure}

		Once all six sets of dictionaries and classifiers are trained the test data was reconstructed using Equation \ref{tst_spcd} and then the classification is done via Equation \ref{TDDL_out}. The results are shown in the ROC curve in Figure \ref{ROC_JMPvTDF}.  From these ROC curves it can be seen that for the data used in this experiment the TD-eFUMI algorithm is able to detect nearly all of the targets found by the detector at a false alarm rate that is significantly lower than the prescreener.

		With this study it is also worth noting that the metal detector will not be capable of detecting every target, as many of them do not contain metal.  This explains why the ROC curves never actually get to 100\% detection.
		
\begin{figure}[thb]
			\centering
			\begin{minipage}{.48\textwidth}
				\centering
				\includegraphics[width=.9\textwidth]{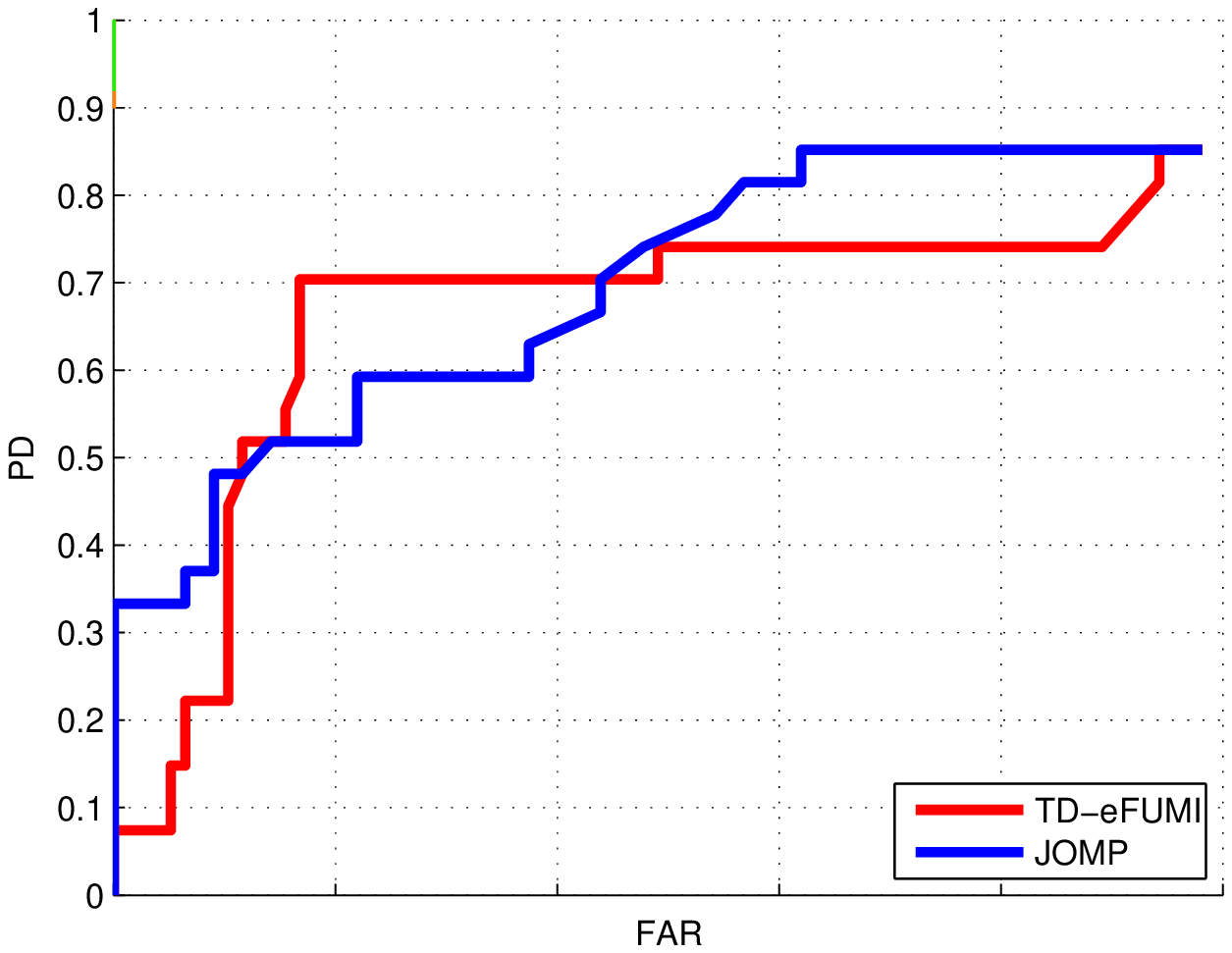}
				\caption{ROC curve for only high metal targets}
				\label{MT_only}
			\end{minipage}
			\begin{minipage}{.48\textwidth}
				\centering
				\includegraphics[width=.9\textwidth]{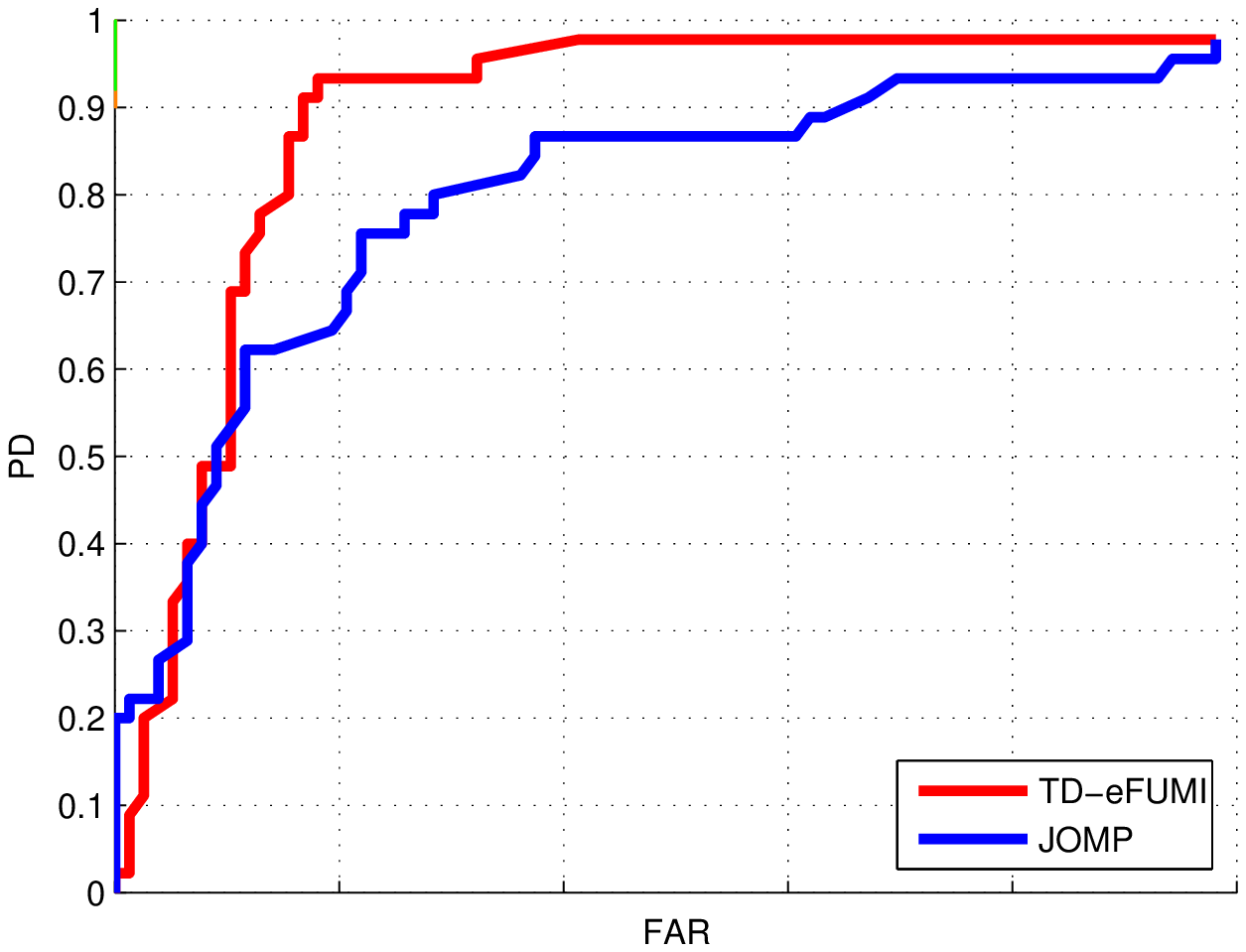}
				\caption{ROC curve for only low metal targets}
				\label{LMT_only}
			\end{minipage}
		\end{figure}

		Two additional ROC curves are shown in Figures \ref{MT_only} and \ref{LMT_only}.  In these ROC curves only a portion of the targets are used.  In Figure \ref{MT_only} the ROC curve is for only the high metal targets and the ROC curve in Figure \ref{LMT_only} is for the low metal targets.  There are many more low metal targets than there are high metal objects so the low metal ROC curve will appear much more similar to the full results in Figure \ref{ROC_JMPvTDF}.  

		The scoring results shown  in this paper were computed while ignoring clutter objects, all other targets are included in the results.  The reason for ignoring clutter during scoring is that it is not anticipated that the proposed algorithm can differentiate between metal targets and metal clutter at this time.  The process of ignoring clutter is to remove any hits that appear close to any of the ground truth locations for the clutter objects within a fixed halo. 
		
		These two ROC curves indicate that the TD-$e$FUMI algorithm does much better than the JOMP prescreener for the low metal targets included in this data set, but yet trails when only considering high metal targets.  As there are many more low metal targets these represent the majority of training samples therefore allowing the TD-$e$FUMI algorithm to learn a better model for the low metal objects versus the high metal ones.  JOMP does not have this problem because it does not require any training which means it will find the objects that do not occur very often better than an algorithm that has to be trained.
		
	\section{Summary and Future Work}
		\label{Summ}
		
			This paper investigated the TD-$e$FUMI algorithm and its effectiveness for landmine detection and classification in EMI data.  Based on the results presented in this paper the proposed TD-$e$FUMI classifier is able to improve upon the JOMP prescreener, however more testing needs to be conducted to make any definitive conclusions.
			
			Potential future work could be to modify the TD-$e$FUMI framework in order to allow for multi-class classification to allow for detecting different types of mines or mine features instead of an all encompassing target class.
		
	\acknowledgments

	This work was funded by Army Research Office grant number 66398-CS to support the US Army RDECOM CERDEC NVESD.  The views and conclusions contained in this document are those of the authors and should not be interpreted as representing the official policies either expressed or implied, of the Army Research Office, Army Research Laboratory, or the U.S. Government. The U.S. Government is authorized to reproduce and distribute reprints for Government purposes notwithstanding any copyright notation hereon.

	\bibliography{refs}
	\bibliographystyle{spiebib}

\end{document}